\documentclass[sigconf]{acmart}

\usepackage{booktabs} % For formal tables
\usepackage{mathrsfs}  
\usepackage{algorithm}
\usepackage{algorithmic}

\makeatletter
\def\BState{\State\hskip-\ALG@thistlm}
\makeatother

\acmDOI{10.475/123_4}

\acmISBN{123-4567-24-567/08/06}

\acmConference[RecSys'18]{ACM Confernce on Recommender Systems}{October 2018}{Vancouver, Canada}
\acmYear{2018}
\copyrightyear{2018}
\acmArticle{4}
\acmPrice{15.00}

\begin{document}
\title{Adversarial Training of Word2Vec \\
	for Basket Completion}

%\author{Anonymous Author}
%\affiliation{
%  \institution{Anonymous Institution}}
%\email{anonymous@anonymous.com}

 \author{Ugo Tanielian}
 \affiliation{%
   \institution{Criteo Research}}
 \affiliation{%
   \institution{UPMC}
   \city{Paris}
   \state{France}}
 \email{u.tanielian@criteo.com}
 
 \author{Mike Gartrell}
 \affiliation{%
   \institution{Criteo Research}
   \city{Paris}
   \state{France}
 }
 \email{m.gartrell@criteo.com}
 
 \author{Flavian Vasile}
 \affiliation{%
   \institution{Criteo Research}
   \city{Paris}
   \state{France}
   \postcode{43017-6221}
 }
 \email{f.vasile@criteo.com}


\begin{abstract}
In recent years, the Word2Vec model trained with the Negative Sampling loss
function has shown state-of-the-art results in a number of machine learning
tasks, including language modeling tasks, such as word analogy and word
similarity, and in recommendation tasks, through Prod2Vec, an extension that
applies to modeling user shopping activity and user preferences. Several methods
that aim to improve upon the standard Negative Sampling loss have been proposed.
In our paper we pursue more sophisticated Negative Sampling, by leveraging ideas
from the field of Generative Adversarial Networks (GANs), and propose
\emph{Adversarial Negative Sampling}.  We build upon the recent progress made in
stabilizing the training objective of GANs in the discrete data setting, and
introduce a new \emph{GAN-Word2Vec} model.  We evaluate our model on the task of
basket completion, and show significant improvements in performance over
Word2Vec trained using standard loss functions, including Noise Contrastive
Estimation and Negative Sampling.
\end{abstract}

%Until recently, Generative Adversarial Networks had not been successfully applied to discrete settings. One of the major reasons lie within the complexity of propagating the gradients back to the generator's weights.



\maketitle

\section{Introduction}
The recommendation task of basket completion is a key part of many online retail
applications. Basket completion involves computing predictions for the next item
that should be added to a shopping basket, given a collection of items that the
user has already added to the basket.

In this context of basket completion, learning item embedding representations
can lead to state-of-the-art results, as shown in~\cite{Deep_Sets}.  Within this
class of approaches, \emph{Word2Vec}~\cite{mikolov2013efficient}, and its
item-based extension \emph{Prod2Vec}~\cite{grbovic2015commerce}, have become the
de-facto standard approach, due to the conceptual simplicity, implementation
simplicity, and state-of-the-art performance of these models.

In terms of training and the use of negatives, there have been many extensions
of the classical \emph{Word2Vec} model based on the \emph{Negative Sampling (NS)
loss function}~\cite{w2v}, such as \emph{Swivel}~\cite{swivel}. However, these
approaches do not have a dynamic way of sampling the most informative negatives.
This shortcoming was addressed by~\cite{chen2018improving}, which proposes
an active sampling heuristic. In our paper we propose \emph{GAN-Word2Vec}, an
extension of Word2Vec that uses ideas from Generative Adversarial Networks
(GANs) to create an \emph{adversarial negative sampling} approach that places
dynamic negative sampling on firmer theoretical grounds, and shows significant
improvements in performance over the classical training approaches.
In our current structure the generator is also trained adversarially and benefits from
a better training signal coming from the discriminator. In terms of training stability, which
becomes an issue in GAN-like settings, our algorithm
builds upon recent advances that make GAN training stable for discrete input
data.  We evaluate the performance of our \emph{GAN-Word2Vec} model on a
basket-completion task and show that it outperforms classical supervised
approaches such as Word2Vec with Negative Sampling by a significant margin.

Overall, the main contributions of this paper are the following:
\begin{itemize}
    \item We propose a new dynamic negative sampling scheme for Word2Vec based
    on ideas from GANs.  To the best of our knowledge, we are the first to
    implement adversarial training for Word2Vec.
    \item We introduce a stable training algorithm that implements our
    adversarial sampling scheme.
    \item Through an experimental evaluation on two real-world datasets, we show
    that our \emph{GAN-Word2Vec} model outperforms classical sampling schemes for
    Word2Vec.
\end{itemize}

We briefly discuss related work on sampling schemes for Word2Vec and the recent
developments on GAN training in discrete settings in Section 2 of this paper. In
Section 3, we formally introduce our \emph{GAN-Word2Vec} model and describe the
training algorithm. We highlight the performance of our method in Section 4, and
conclude with main ideas and directions for future work in Section 5.

\section{Related Work}

\subsection{Basket Completion with Embedding Representations}
In the recent years, a substantial amount of work has focused on improving the
performance of language modeling and text generation.  In both tasks, Deep
Neural Networks (DNNs) have proven to be extremely effective, and are now
considered state-of-the-art.  In this paper, we focus on the task of basket
completion with learned embedding representations. For the task of basket
completion, very little work has focused on applying DNNs, with the notable
exception of~\cite{Deep_Sets}. In this paper, the authors introduced a new
family of networks designed to work on sets, where the output is invariant to
any permutation in the order of objects in the input set.  As an alternative to
a set-based interpretation, basket completion can be approached as a sequence
generation task, where one seeks to predict the distribution of the next item
conditioned on the items already present in the basket. First, one could use the
Skip-Gram model proposed by~\cite{w2v}, and use the average of the embeddings
for the items within a basket to compute the next-item prediction for the
basket. Second, building upon work on models for text generation, it is natural to
leverage Recurrent Neural Networks (RNNs), particularly Long Short Term Memory cells,
as proposed by~\cite{lstm}, or bi-directional
LSTMs~\cite{bidirectionalLSTMs,bidirectionalRNNs}. Convolutional neural networks
could also be used, for example by using a \textit{Text-CNN} like architecture
as proposed by~\cite{TextCNN}. The authors of~\cite{TextCNN} empirically show on
different tasks, such as point cloud classification, that their method
outperforms state-of-the-art results with good generalization properties.

%\begin{itemize}
%\item Avg(Word2Vec)
%\item TextCNN
%\item DeepSets
%\item RNNs?
%\end{itemize}

\subsection{Negative sampling schemes for Word2Vec}
When dealing with the training of multi-class models with thousands or millions
of output classes, candidate sampling algorithms can speed up the training by
considering a small randomly-chosen subset of contrastive candidates for each
batch of training examples.  Ref.~\cite{NCE_loss} introduced Noise Contrastive
Estimation (NCE) as an unbiased estimator of the softmax loss, and has been
proven to be efficient for learning word embeddings~\cite{mnih2012fast}. In
~\cite{w2v}, the authors propose \textit{negative sampling} and directly sample
candidates from a noise distribution.

More recently, in~\cite{chen2018improving}, the authors provide an insightful
analysis of negative sampling. They show that negative samples with high inner
product scores with a context word are more informative in terms of gradients on
the loss function.  Leveraging this analysis, the authors propose a dynamic
sampling method based on inner-product rankings. This result can be intuitively
interpreted by seeing that negative samples with a higher inner product will
lead to a better approximation of the softmax.

In our setup, we simultaneously train two neural networks, and use the output
distribution coming from one network to generate the negative samples for the
second network. This \textit{adversarial negative sampling} proves to be a dynamic and
efficient way to improve the training of our system. This method echoes the
recent work on Generative Adversarial Networks.

%\begin{itemize}
%\item NS
%\item Glove
%\item Swivel
%\item Other Missing Not at Random
%\end{itemize}

\subsection{GANs}
First proposed by~\cite{GANs} in 2014, Generative Adversarial Networks (GANs)
have been quite successful at generating realistic images.  GANs can be viewed
as a framework for training generative models by posing the training procedure
as a minimax game between the generative and discriminative models.  This
adversarial learning framework bypasses many of the difficulties associated with
maximum likelihood estimation (MLE), and has demonstrated impressive results in
natural image generation tasks.

Theoretical work~ ~\cite{LiBoCh07,ZhLiZhXuHe18,ArBo17,NoCsTo16} and practical
studies~\cite{SaGoZaChRaCg16,ArChBo17,DzRoGh15} have stabilized GAN training and
enabled many improvements in continuous spaces. However, in discrete settings
a number of important limitations have prevented major breakthroughs.
A major issue involves the complexity of backpropagating the gradients for the
generative model. To bypass this differentiation
problem,~\cite{SeqGANs,MALIGANs} proposed a cost function inspired by
reinforcement learning. The discriminator is used as a reward function and the
generator is trained via policy gradient~\cite{sutton2000policy}. Additionally,
the authors propose the use of importance sampling and variance reduction
techniques to stabilize the training.

%\begin{itemize}
%\item GAN-classical
%\item MaliGAN
%\end{itemize}

\section{Our Model: GAN-Word2Vec}
We begin this section by formally defining the basket completion task. We denote
$\mathbb{Z}$ as all of the potential items, products, or words. $\mathbb Z^{d}$
is the space of all baskets of size $d$, and $U = \bigcup_{d=1}^{\infty} \mathbb
Z^d$ is the space of all possible baskets of any size.

The objective of GANs is to generate candidate, or synthetic, samples from a target
data distribution $p^{\star}$, from which only true samples -- in our case
baskets of products -- are available. Therefore, the collection of true baskets
$S$ is a subset of $U$. We are given each basket $\{X\} \in S$, and an element $z
\in Z$ randomly selected from $\{X\}$.  Knowing $\{X \backslash z\}$, which
denotes basket $X$ with item $z$ removed, we want to be able to predict the
missing item $z$. We denote $\mathbb{P(Z)}$ as the space of probability
distributions defined on $\mathbb{Z}$.

As in~\cite{w2v}, we are working with embeddings. For a given target item $z$,
we denote $C_z$ as its context, where $C_z = \{X \backslash z \}$. The
embeddings of $z$ and $C_z$ are $w_z$ and $w_{C_z}$, respectively.

\subsection{Notation}

Both generators and discriminators as a whole have the form of a parametric
family of functions from $\mathbb Z^{d'}$ to $\mathbb{P(Z)}$, where $d'$ is the
length of the prefix basket.  We denote $\mathscr G=\{G_{\theta}\}_{\theta \in
\Theta}$ as the family of generator functions, with parameters $\Theta \subset
\mathbb R^p$, and $\{D_{\alpha}\}_{\alpha \in \Lambda}$ as the family of
discriminator functions, with parameters $\Lambda \subset \mathbb R^q$. Each
function $G_{\theta}$ and $D_{\alpha}$ is intended to be applied to a $d'$-long
random input basket. Both the generator and the discriminator output
probabilities that are conditioned on an input basket. In our setting, $G$ and $D$
may therefore be from the same class of models, and may be parametrized by the
same class of functions. Given the context $C_z$, we denote
$P_G(.|C_z)$ and $P_D(.|C_z)$ as the conditional probabilities output by $G$ and
$D$, respectively. The samples drawn from $P_G(.|C_z)$ are denoted as $O$.

\subsection{Adversarial Negative Sampling Loss}

In the usual GAN setting, $D$ is trained with a binary cross-entropy loss
function, with positives coming from $p^{\star}$ and negatives generated by $G$.
In our setting, we modify the discriminator's loss using an extension of the
standard approach to \emph{Negative Sampling}, which has proven to be very
efficient for language modeling. We define \textit{Adversarial Negative
Sampling Loss} by the following objective:

\begin{equation} \label{eq:1}
\log \ \sigma (w_z \ w_{C_z}^T) + \sum_{i=1}^{k} \ \mathbb{E}_{O_i \sim P_G(.|C_z)} \ [\log \ \sigma (-w_{O_i} \ w_{C_z}^T)]
\end{equation}
where $k$ is the number of negatives $O_i$ sampled from $P_G(.|C_z)$.
%\begin{equation}
%log \ \sigma (v'_{w_z} v_{w_I}^T) + \sum_{i=1}^{k} \ \mathbb{E}_{w_i \sim P_G(w_I)} \ [log \ \sigma (-%v'_{w_i} v_{w_I}^T)]
%\end{equation}

As in the standard GAN setting, D's task is to learn to discriminate between
true samples and synthetic samples coming from $G$. Compared to standard
\textit{negative sampling}, the main difference is that the negatives are now
drawn from a dynamic distribution that is by design more informative than the
fixed distribution used in standard negative sampling.
Ref.~\cite{chen2018improving} proposes a dynamic sampling strategy based on
self-embedded features, but our approach is a fully adversarial sampling method
that is not based on heuristics.

\subsection{Training the Generator}
In discrete settings, training $G$ using the standard GAN architecture is not
feasible, due to discontinuities in the space of discrete data that prevent
updates of the generator. To address this issue, we sample potential outcomes
from $P_G(.|C_z)$, and use $D$ as a reward function on these outcomes.
%as one cannot calculate $ \frac{\partial D_{\alpha}(G_{\theta})}{\partial
%\theta}$.

In our case, the training of the generator has been inspired
by~\cite{MALIGANs}. In this paper, the authors define two main loss functions for
training the generator. The first loss is called the \textit{basic MALIGAN}
and only uses signals from $D$. Adapted to our setting, we have the
following formulation for the generator's loss:

\begin{equation} \label{eq:2}
L_G(\theta) = \sum_{i=1}^{m} \ \frac{r_D(O_i)}{\sum_i{r_D(O_i)}} \ \log \ P_G(\mathit{O_i}|C_z)
\end{equation}

where:
\begin{itemize}
\item $O_i \sim P_G(.|C_z)$.  That is, we draw negative samples in the form of
``next-item'' samples, where only the missing element $z$ for each basket
$\{X\}$ is sampled.  This missing element is sampled by conditioning $P_G$ on
the context, $C_z$, for $z$.
\item $r_D(O_i) = \frac{p_D(\mathit{O_i}|C_z)}{1 \ - \ p_D(\mathit{O_i}|C_z})$.
This term allows us to incorporate reward from the discriminator into updates
for the generator.  The expression for $r_D$ comes from a property of the
optimal $D$ for a GAN; a full explanation is provided in~\cite{MALIGANs}.
\item $m$ is the number of negatives sampled from $P_G(.|C_z)$ used to compute
the gradients
\end{itemize}

Unlike~\cite{MALIGANs}, we do not use any $b$ parameter in Eq.~\ref{eq:2}, as
we did not observe the need for further variance reduction provided by this term
in our applications. As both models output probability distributions, computing
$P_G(\mathit{O_i}|C_z)$ and $r_D(O_i)$ is straightforward.

The second loss, \textit{Mixed MLE-MALIGAN}, mixes adversarial loss and the standard
maximum likelihood estimate (MLE) loss. In our case, this loss mixes the
adversarial training loss and a standard sampled softmax loss (negative sampling
loss):

\begin{align} \label{eq:3}
L_G(\theta) &= 0.5 \ * \ \sum_{i=1}^{m} \ \frac{r_D(O_i)}{\sum_i{r_D(O_i)}} \ \log \ P_G(\mathit{O_i}|C_z) \ \ +  \nonumber \\
& 0.5 \left( \log \ \sigma (w_z \ w_{C_z}^T) + \sum_{i=1}^{k} \ \mathbb{E}_{N_i \sim U(Z)} \ [\log \ \sigma (-w_{N_i} \ w_{C_z}^T)] \right)
\end{align}
where $N_i \sim U(Z)$ are negatives uniformly sampled among the potential next
items.  We empirically find that this mixed loss provides more stable gradients
than the loss in Eq.~\ref{eq:2}, leading to faster convergence during training.

%Using this adversarial loss could be problematic in the case of mode collapsing. Indeed, if $G$ collapses to one mode, then $P_G(.|C_z)$ will have very few entropy and and the signal $r_D(O_i)$ where $O_i$ is sampled from $P_G(.|C_z)$ will be noisy because discriminating elements from this mode only was not the task $D$ was trained on. Therefore, we encourage to add some noise in the sampling of this $O_i$ in computing $G$'s loss. We propose the following formulation:
%\begin{align} \label{eq:4}
%L_G(\theta) = &\sum_{i=1}^{m} \ \frac{r_D(O_i)}{\sum_i{r_D(O_i)} + \sum_k{r_D(N_k)}} \ \log %\ P_G(\mathit{O_i}|C_z) \ +  \nonumber \\
%&\sum_{k=1}^{q} \ \frac{r_D(N_k)}{\sum_i{r_D(O_i)} + \sum_k{r_D(N_k)}} \ \log \ P_G(\mathit{N_k}|C_z)
%\end{align}
%where the negatives $N_i$ uniformly sampled are directly mixed with the $O_i$ sampled from %$P_G(\mathit{O_i}|C_z)$. By doing so, gradients from $G$ get more stabilized. in practice, we found out that using $m=2*q$ was a good ratio. 

%\subsection{Algorithm}
A description of our algorithm can be found in Algorithm \ref{alg:framework}. We
pre-train both $G$ and $D$ using a standard negative sampling loss before
training these components adversarially. We empirically show improvements with
this procedure in the following section.

\begin{algorithm}[t]
\caption{GAN-Word2Vec}\label{alg:framework}
\begin{algorithmic}
\small
\REQUIRE
generator policy $G_\theta$; discriminator $D_\alpha$; a basket dataset $\mathcal{S}$

\STATE
Initialize $G_\theta$, $D_\alpha$ with random weights $\theta,\alpha$.
\STATE
Pre-train $G_\theta$ and $D_\alpha$ using sampled softmax on $\mathcal{S}$

\REPEAT
\FOR{g-steps}
\STATE
Get random true sub-baskets $\{X \backslash z \}$ and the targets $z$.
\STATE
Generate negatives by sampling from $P_G(.|\{X \backslash z \})$
\STATE
Update generator parameters via policy gradient Eq.~(\ref{eq:3})
\ENDFOR

\FOR{d-steps}
\STATE
Get random true sub-baskets $\{X \backslash z \}$ and the targets $z$.
\STATE
Generate adversarial samples for $D$ from $P_G(.|\{X \backslash z \})$
\STATE
Train discriminator $D_\alpha$ by Eq.~({\ref{eq:1}})
\ENDFOR
\UNTIL{GAN-Word2Vec converges}
\end{algorithmic}
\end{algorithm}

\section{Experiments}

All our experiments have been ran on the task basket completion, which is a well-known Recommendation task. \

\subsection{Datasets}
In~\citep{gartrell2016bayesian} and~\citep{gartrell2017low}, the authors present
state-of-the-art results on basket completion datasets. We performed our
experiments on two of the datasets used in this prior work: the \textit{Amazon Baby
Registries} and the \textit{Belgian Retail} datasets.
\begin{itemize}
\item This public dataset consists of registries of baby products from 15
different categories (such as 'feeding', 'diapers', 'toys', etc.), where the
item catalog and registries for each category are disjoint. Each category
therefore provides a small dataset, with a maximum of 15,000 purchased baskets
per category. We use a random split of 80\% of the data for training and 20\%
for testing.

\item Belgian Retail Supermarket - This is a public dataset composed of shopping
baskets purchased over three non-consecutive time periods from a Belgian retail
supermarket store. There are 88,163 purchased baskets, with a catalog of 16,470 unique
items. We use a random split of 80\% of the data for training and 20\% for
testing.
\end{itemize}

\subsection{Task definition and associated metrics}
In the following evaluation, we consider two metrics:
\begin{itemize}
\item Mean Percentile Rank (MPR) - For a basket $\{X\}$ and one item $z$
randomly removed from this basket, we rank all potential items $i$ from set of candidates
$\mathbb{Z}$ according to their probabilities of completing $\{X \backslash
z \}$, which are $P_G(i | \{X \backslash z \})$ and $P_D(i|\{X \backslash z \})$.
The Percentile Rank (PR) of the missing item $z$ is defined by:
\begin{equation}
\text{PR}_j = \frac{\sum_{j \prime \in \mathbb{Z}}\mathbb{I}\Large( p_j \ge p_{j \prime}\Large)}{|\mathbb{Z}|} \times 100\% \nonumber
\end{equation}

where $\mathbb{I}$ is the indicator function and $|\mathbb{Z}|$ is the number of items in the
candidate set. The Mean Percentile Rank (MPR) is the average PR of all the
instances in the test-set $\mathcal{T}$.
\begin{equation}
\text{MPR} = \frac{\sum_{t \in \mathcal{T}} \text{PR}_t}{|\mathcal{T}|} \nonumber
\end{equation}
MPR = 100 always places the held-out item for the test instance at the head of
the ranked list of predictions, while MPR = 50 is equivalent to random
selection.

\item Precision@k - We define this metric as
\begin{equation}
    \text{precision@}k = \frac{\sum_{t \in \mathcal{T}} \mathbb{I}[\text{rank}_t \leq k]}{|\mathcal{T}|} \nonumber
\end{equation}
where $rank_t$ is the predicted rank of the held-out item for test instance $t$.
In other words, precision@k is the fraction of instances in the test set for
which the predicted rank of the held-out item falls within the top $k$
predictions.

\end{itemize}

\subsection{Experimental results}

We compare our \emph{GAN-Word2Vec} model with Word2Vec models training using
classical loss funcitions, including \emph{Noise Contrastive Estimation Loss}
(NCE)~\cite{NCE_loss,mnih2012fast} and \emph{Negative Sampling Loss} (NEG) ~\cite{w2v}. We
observe that we have better results with the \textit{Mixed Loss}.  

We find that pre-training both $G$ and $D$ with a \emph{Negative Sampling Loss} leads to better predictive quality for \emph{GAN-Word2Vec}. 

After pre-training, we train $G$ and $D$ using Eq.~\ref{eq:3} and Eq.~\ref{eq:1}, respectively. 
We observe that the discriminator initially benefits from adversarial sampling, 
and its performance on both MPR and precision@1 increases. However, after
convergence, the generator ultimately provides better performance than the
discriminator on both metrics. We conjecture that this may be explained 
by the fact that basket completion is a generative task.

From Table~\ref{tab:final_results_belgian}, we see that our GAN-Word2Vec model
consistently provides statistically-significant improvements over the Word2Vec baseline
models on both the Precision@1 and MPR metrics. As confirmed by the experiments,
we expect our method to be more effective on larger datasets.

We also see that Word2Vec trained using Negative Sampling (W2V-NEG) is generally
a stronger baseline than Word2Vec trained via NCE. 

%Since our GAN-Word2Vec model builds on Word2Vec with negative sampling (W2V-NEG), 
%we conclude that our adversarial approach leads to
%predictive quality that cannot be achieved through standard approaches to
%training Word2Vec that we considered here.

\begin{table}
\centering
\begin{tabular}{l|c|c}
Method & Precision@1 & MPR \\
\hline\hline
Amazon dataset \\
\hline
W2V-NCE & 14.80 $\pm0.07$& 80.15 $\pm0.05$\\
W2V-NEG & 15.40 $\pm0.05$& 80.20 $\pm0.07$\\
W2V-GANs & \textbf{\textit{16.30} $\pm0.08$}& \textbf{\textit{80.50} $\pm0.1$}\\
\\
\hline
Belgian retail dataset \\
\hline
W2V-NCE & 29.50 $\pm0.05$& 87.54 $\pm0.04$\\
W2V-NEG & 34.35 $\pm0.07$& 88.55 $\pm0.05$\\
W2V-GANs & \textbf{\textit{35.82} $\pm0.09$}& \textbf{\textit{89.45} $\pm0.1$}\\
\end{tabular}
\caption{One item basket completion task on the Belgian retail dataset.}
\label{tab:final_results_belgian}
\end{table}

\section{Conclusions}

In this paper, we have proposed a new adversarial negative sampling algorithm
suitable for models such as Word2Vec. Based on recent progress made on GANs in
discrete data settings, our solution eliminates much of the complexity of
implementing a generative adversarial structure for such models. In particular,
our adversarial training approach can be easily applied to models that use
standard sampled softmax training, where the generator and discriminator can be
of the same family of models.

Regarding future work, we plan to investigate the effectiveness of this training
procedure on other models. It is possible that models with more capacity
than \textit{Word2Vec} could benefit even more from using softmax
with the adversarial negative sampling loss structure that we have proposed.
Therefore, we plan to test this procedure on models such as TextCNN, RNNs, and
determinantal point processes (DPPs)~\cite{gartrell2016bayesian,gartrell2017low},
which are known to be effective in modeling discrete set structures.

GANs have proven to be quite effective in conjunction with deep neural networks
when applied to image generation.  In this work, we have showed that adversarial
training can also be applied to simpler models, in discrete settings, and
bring statistically significant improvements in predictive quality.

\bibliographystyle{ACM-Reference-Format}
\bibliography{adversarial_training}

\end{document}